\documentclass[letterpaper, 10pt, conference]{ieeeconf}      

\IEEEoverridecommandlockouts                  

\usepackage{graphicx}
\usepackage{xcolor}
\usepackage{hyperref}
\usepackage{multirow}

\title{\LARGE \bf MultiPhysio-HRC: Multimodal Physiological Signals Dataset for industrial Human-Robot Collaboration}

\author{Andrea Bussolan$^{1}$, Stefano Baraldo$^{1}$, Oliver Avram$^{1}$, Pablo Urcola$^{3}$, Luis Montesano$^{3, 4}$, \\ Luca Maria Gambardella$^{2}$, and Anna Valente$^{1}$
\thanks{*The research in this paper has been partially funded by the Horizon Europe project Fluently (Grant ID: 101058680) and Eurostars project !2309-Singularity.}
\thanks{$^{1}$Andrea Bussolan, Stefano Baraldo, Oliver Avram and Anna Valente are with the ARM-Lab, SUPSI, Lugano, 6900, Switzerland {\tt\small \{andrea.bussolan, stefano.baraldo, oliver.avram, anna.valente\}@supsi.ch}}%
\thanks{$^{2}$Luca Maria Gambardella is with the Faculty of Informatics, USI, Lugano, 6900, Switzerland {\tt\small luca.gambardella@usi.ch}}%
\thanks{$^{3}$Pablo Urcola and Luis Montesano are with Bitbrain, Zaragoza, Spain{\tt\small \{pablo.urcola, luis.montesano\}@bitbrain.es}}%
\thanks{$^{3}$Luis Montesano is with the Universidad de Zaragoza, Zaragoza, Spain {\tt\small lmontesa@unizar.es}}%
}

\begin{document}

\maketitle
\thispagestyle{empty}
\pagestyle{empty}

\begin{abstract}
Human-robot collaboration (HRC) is a key focus of Industry 5.0, aiming to enhance worker productivity while ensuring well-being. The ability to perceive human psycho-physical states, such as stress and cognitive load, is crucial for adaptive and human-aware robotics. This paper introduces MultiPhysio-HRC, a multimodal dataset containing physiological, audio, and facial data collected during real-world HRC scenarios. The dataset includes electroencephalography (EEG), electrocardiography (ECG), electrodermal activity (EDA), respiration (RESP), electromyography (EMG), voice recordings, and facial action units. The dataset integrates controlled cognitive tasks, immersive virtual reality experiences, and industrial disassembly activities performed manually and with robotic assistance, to capture a holistic view of the participants' mental states. Rich ground truth annotations were obtained using validated psychological self-assessment questionnaires. Baseline models were evaluated for stress and cognitive load classification, demonstrating the dataset’s potential for affective computing and human-aware robotics research. MultiPhysio-HRC is publicly available to support research in human-centered automation, workplace well-being, and intelligent robotic systems.
\end{abstract}

\section{Introduction}

In the field of Human-Robot Collaboration (HRC), physiological signals are raising high interest thanks to their potential to capture human states such as stress, cognitive load, and fatigue \cite{lorenzini2023}. In the human-centric view promoted by Industry 5.0, industrial workplaces should aim at striking a balance between worker productivity and well-being \cite{lu2022}. This includes conceiving robotic systems that can not only perform physical tasks in support of human workers but also change their behavior depending on the psycho-physical state of operators, coupled with context information. This approach of \textit{deliberative robotics} \cite{valente2022} cannot unleash its full potential unless the human psycho-physical state can be perceived by the robot. This idea is the core goal of the \textit{Fluently} project, which aims to enhance human-robot collaboration by enabling robots to adapt their behavior based on the psycho-physical state of human operators.

To develop robotic systems capable of adapting to human states, it is essential to build machine learning models that can reliably infer the mental state from physiological and behavioral signals. However, training such models requires datasets that not only include a diverse range of conditions but also reflect real-world industrial settings. Many existing datasets focus on a limited subset of modalities and are rarely collected outside of controlled laboratory conditions, limiting their applicability to HRC scenarios where multiple factors influence human states simultaneously.

In this paper, we present \textbf{MultiPhysio-HRC}, a dataset containing facial features, audio, and physiological signals - electrocardiogram (ECG), electrodermal activity (EDA), respiration (RESP), electromyography (EMG), and Electroencephalography (EEG). To the best of our knowledge, \textbf{MultiPhysio-HRC} is the first dataset to include this wide combination of data obtained during real-world human-robot collaboration, various psychological tests, and VR-based activities, designed to elicit multiple psychological states. Furthermore, the ground truth labels collected for this dataset enable the analysis of various aspects of the human mental state, including stress levels, cognitive load, and emotional dimensions. The dataset is publicly available at \url{https://automation-robotics-machines.github.io/MultiPhysio-HRC.github.io/}.

We summarize our main contributions as follows:

\begin{itemize}
    \item \textbf{Real-World HRC Context} - To the best of our knowledge, MultiPhysio-HRC is the first publicly available dataset to include real-world industrial-like HRC scenarios comprehensively.
    \item \textbf{Complete Multimodal Data} - While existing datasets often include subsets of modalities, MultiPhysio-HRC integrates facial features, audio, and a comprehensive set of physiological signals: EEG, ECG, EDA, RESP, and EMG. This combination allows for a holistic assessment of mental states, addressing cognitive load, stress, and emotional dimensions.
    \item \textbf{Task Diversification} - The dataset comprises tasks specifically designed to elicit various mental states. These include cognitive tests, immersive VR activities, and industrial tasks.
    \item \textbf{Rich Ground Truth Annotations} - Ground truth labels were collected through validated psychological questionnaires at multiple stages during the experiment. Combined with multimodal measurements, these labels offer unparalleled granularity for studying human states in HRC contexts.
\end{itemize}

\begin{figure*}[t!]
    \centering
    \includegraphics[width=\linewidth]{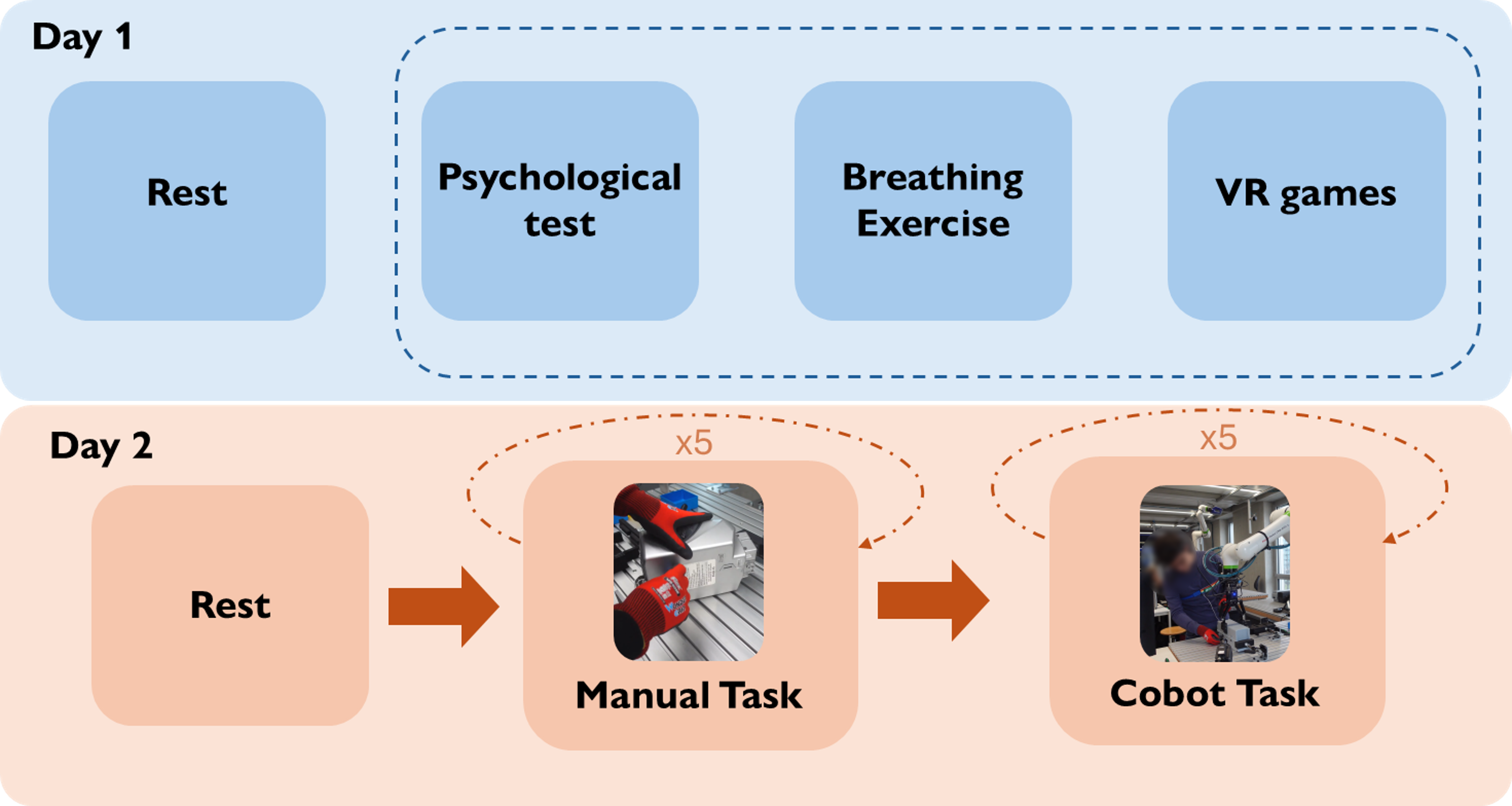}
    \caption{Data acquisition protocol.}
    \label{fig: protocol}
\end{figure*}

The remainder of this paper is organized as follows: Section \ref{sec: sota} presents the related dataset with similar modalities combination; Section \ref{sec: dataset} explains the experimental protocol for data collection, describing tasks and data; in Section \ref{sec: methods} the processing pipelines for filtering and feature extraction are detailed; while Section \ref{sec: results} presents and discusses the results achieved using traditional models. In the end, Section \ref{sec: conclusion} concludes the work by presenting final remarks and future directions.

\section{Related Works} \label{sec: sota}
The field of \textit{Affective Computing} has a long history of public datasets for emotion and mental state recognition through diverse experimental setups and various physiological and behavioral data combinations. 

One of the first publicly available datasets was published in \cite{healey_detecting_2005}. This dataset features ECG, EDA, RESP, and EMG data on driver stress during real-world driving tasks. The WESAD dataset \cite{schmidt_introducing_2018} is a multimodal dataset aimed at stress and affect detection using wearable sensors. It includes physiological and motion data from 15 participants recorded via both wrist-worn (Empatica E4) and chest-worn (RespiBAN) devices. Sensor modalities include ECG, EDA, EMG, respiration, temperature, and acceleration. Participants were exposed to neutral, stress (via the Trier Social Stress Test), and amusement conditions. Ground truth was collected using PANAS, SAM, STAI, and SSSQ questionnaires. The dataset enables benchmarking of affective state classification with a focus on wearable technology.
The DREAMER dataset \cite{katsigiannis_dreamer_2018} focuses on emotion recognition in response to audiovisual stimuli. It consists of EEG and ECG signals from 23 subjects exposed to 18 short emotional video clips. After each clip, participants self-assessed their emotional state in terms of valence, arousal, and dominance using the SAM (Self-Assessment Manikins) scale. The recordings were collected using low-cost, wireless devices, making the dataset particularly suitable for developing lightweight emotion recognition systems.
In \cite{sarkar_avcaffe_2023}, AVCAffe, a large-scale audio-visual dataset that studies cognitive load and affect in remote work scenarios, is presented. This dataset includes data from 106 participants performing seven tasks via video conferencing. Tasks included open discussions and collaborative decision-making exercises, designed to elicit varying levels of cognitive load. AVCAffe includes annotations for arousal, valence, and cognitive load attributes. StressID \cite{chaptoukaev_stressid_2023} is a comprehensive multimodal dataset specifically designed for stress identification, containing synchronized recordings of facial expressions, audio, and physiological signals (ECG, EDA, respiration) from 65 participants. The dataset features annotated data collected during 11 tasks, including guided breathing, emotional video clips, cognitive tasks, and public speaking scenarios.

However, the number of public datasets focusing on the physiological response of individuals during real-world HRC tasks is extremely limited. The SenseCobot dataset stands out as a structured effort to investigate operator stress during collaborative robot programming tasks \cite{sensecobot_sensecobot_2023, borghi_assessing_2025}. In this study, users were trained to program a UR10e cobot in a simulated industrial setup. The authors collected EEG, ECG, GSR, and facial expressions as input data and used NASA-TLX as ground truth labels. The SenseCobot dataset lacks exposure to complex, task-integrated HRC contexts such as physical collaboration or time-constrained industrial procedures. In contrast, the MultiPhysio-HRC dataset addresses this gap by incorporating a broader range of scenarios, including manual and robot-assisted battery disassembly, cognitive load induction through psychological tests (e.g., Stroop, N-back), and immersive virtual reality tasks. Moreover, MultiPhysio-HRC features a richer set of modalities—including EEG, ECG, EDA, EMG, respiration (RESP), facial action units, and audio features, together with detailed ground truth from validated self-assessment questionnaires (STAI-Y1, NASA-TLX, SAM, and NARS), enabling a more holistic assessment of stress, cognitive load, and emotional state in realistic industrial HRC settings.

\section{\textbf{MultiPhysio-HRC}} \label{sec: dataset}
\subsection{Experimental Protocol} \label{sec: protocol}

The data collection campaign was designed to build a multimodal and multi-scenario dataset for mental state assessment, integrating psychological, physiological, and behavioral data. The protocol designed for this dataset acquisition is inspired by the work presented in  \cite{buss2023}. The protocol spans two days of activities, focusing on varying stress levels and operational conditions, including human-robot collaboration and manual tasks. A schematic representation of the overall protocol is represented in Fig. \ref{fig: protocol}.

\subsubsection{Day 1 - Baseline and Stress Induction}

\begin{figure}[t]
    \centering
    \includegraphics[width=0.9\columnwidth]{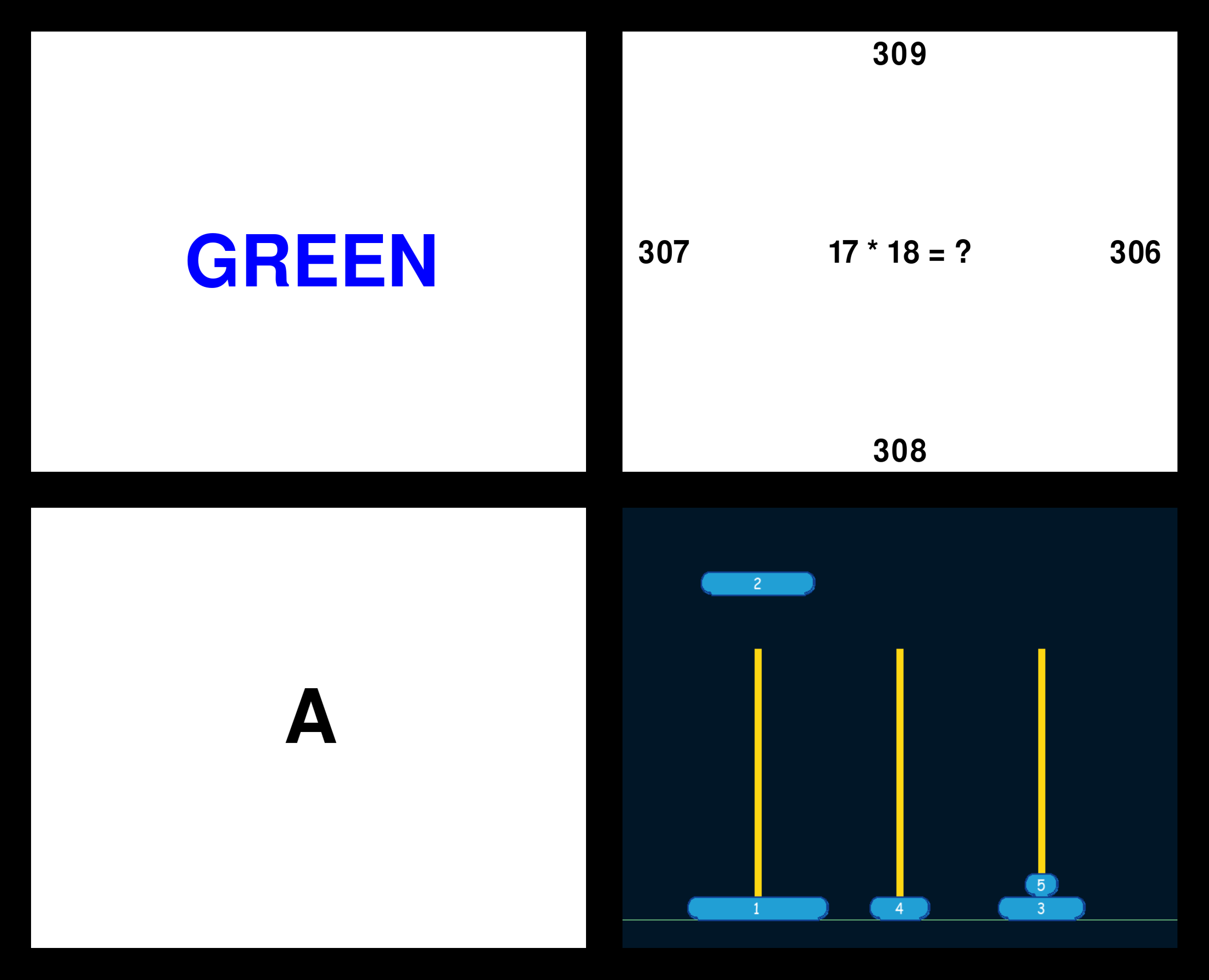}
    \caption{Displayed screen of each cognitive task: SCWT (top left), N-Back (bottom left), Arithmetic (top right), and Hanoi tower (bottom right).}
    \label{fig: screens}
\end{figure}

Participants began with a resting period to establish baseline physiological measures. Following this, they were asked to perform activities including cognitive load tests, breathing exercises, and VR games. In detail:
\begin{enumerate}
    \item \textbf{Rest}. The participant sits comfortably for two minutes and is invited to relax without specific instructions.
    \item \textbf{Cognitive tasks}. The participant sits in front of a computer screen, using a keyboard and mouse to interact with different games aimed at increasing their cognitive load and eliciting psychological stress. The selected tasks are:
    \begin{enumerate}
        \item Stroop Color Word Test (SCWT) \cite{stroop2017} (three minutes). Color names (e.g., "RED") appear in different colors. The participants must push the keyboard button corresponding to the color of the displayed letters (e.g., "B" if the word "RED" is written in Blue characters). The task was performed with two difficulty levels: one second and half a second to answer.
        \item N-Back task \cite{nback2017} (two minutes). A single letter is shown on the screen every two seconds. The participant must press a key whenever the letter is equal to the \textit{N-th} previous letter.
        \item Mental Arithmetic Task (two minutes). The participant must perform a mental calculation in three seconds and press an arrow key, selecting the correct answer among four possibilities.
        \item Hanoi Tower \cite{hanoi2010}. The participant must rebuild the tower in another bin, without placing a larger block over a smaller one. There was no time constraint on this task.
        \item Breathing exercise (two minutes). A voice-guided controlled breathing exercise.
    \end{enumerate}
    The order of these tasks was randomly chosen for each participant. A representation of the displayed screen is shown in Fig. \ref{fig: screens}. During the execution of these tasks (except the Hanoi tower and the breathing exercise), a ticking clock sound was reproduced to arouse a sense of hurry, and a buzzer sound was played in case of mistakes, to increase the psychological stress.
    \item \textbf{VR games}. Finally, participants performed immersive tasks in virtual reality environments such as \textit{Richie's Plank Experience} \footnote{https:$//$store.steampowered.com$/$app$/$517160$/$Richies\_Plank\_Experience} to elicit a high-intensity psycho-physical state. In this game, participants had to walk on a bench suspended on top of a building. 
\end{enumerate}

After each one of these tasks, the ground truth questionnaires were administered (see sec. \ref{sub: ground}).

\subsubsection{Day 2 - Manual and Robot-Assisted Tasks} \label{sec: day2}

The second day was dedicated to a battery disassembly task (described in sec. \ref{sub: task}), designed to compare the experience of fully manual work with HRC. In detail, the second day was structured in the following phases:

\begin{enumerate}
    \item \textbf{Rest}. The participant sits comfortably for five minutes and is invited to relax without specific instructions.
    \item \textbf{Manual disassembly}. The participant uses bare hands or simple tools to partially disassemble an e-bike battery pack.
    \item \textbf{Collaborative disassembly}. The participant is given instructions about how to interact with the robot by voice commands. Then, they perform the same disassembly by asking the cobot to perform support or parallel operations. The voice commands are not only used to give instructions to the robot naturally, but are also opportunities to collect voice data and observe human-robot dynamics under operational conditions.
\end{enumerate}
Each task (manual and robot-assisted) was repeated up to five times to elicit fatigue. After each one of these tasks, the ground truth data was collected.

\begin{figure}[t]
    \centering
    \includegraphics[width=\columnwidth, keepaspectratio]{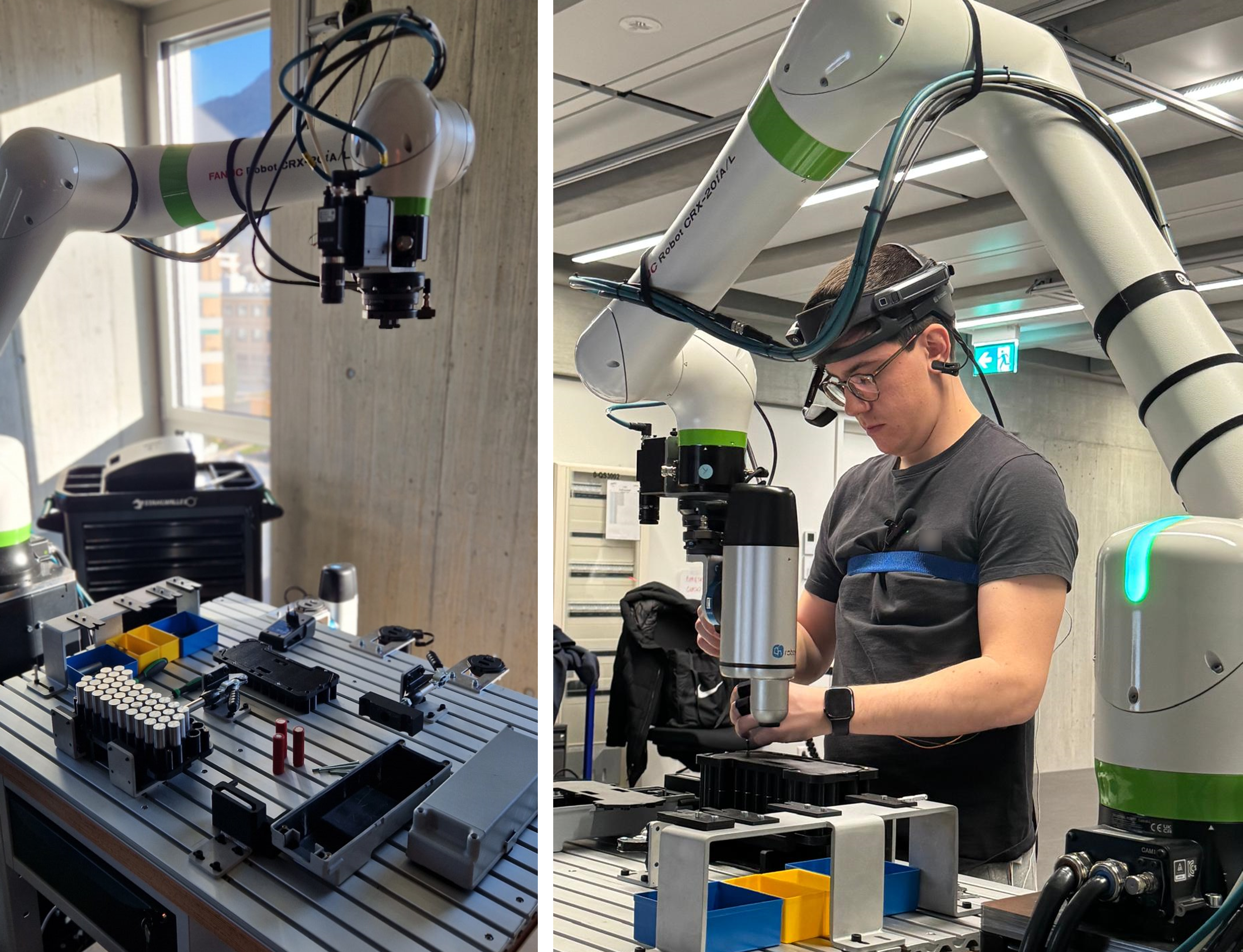}
    \caption{Experimental robotic cell setup. The multiple components of the disassembled battery can be seen placed on the table. }
    \label{fig: robotic-cell}
\end{figure}

\subsection{Task and Robotic Cell Description}\label{sub: task}
The industrial task described in \ref{sec: day2} involves e-bike battery disassembly, a task selected due to its fundamental importance for fostering sustainable industrial practices. Participants performed both manual and collaborative disassembly of various battery models, with procedures designed to adhere to real-world conditions safely. For safety reasons, the original battery cells were replaced with aluminum cylinders of the same shape and dimensions, eliminating soldering materials and hazardous components.

\begin{figure}
    \centering
    \includegraphics[width=\linewidth]{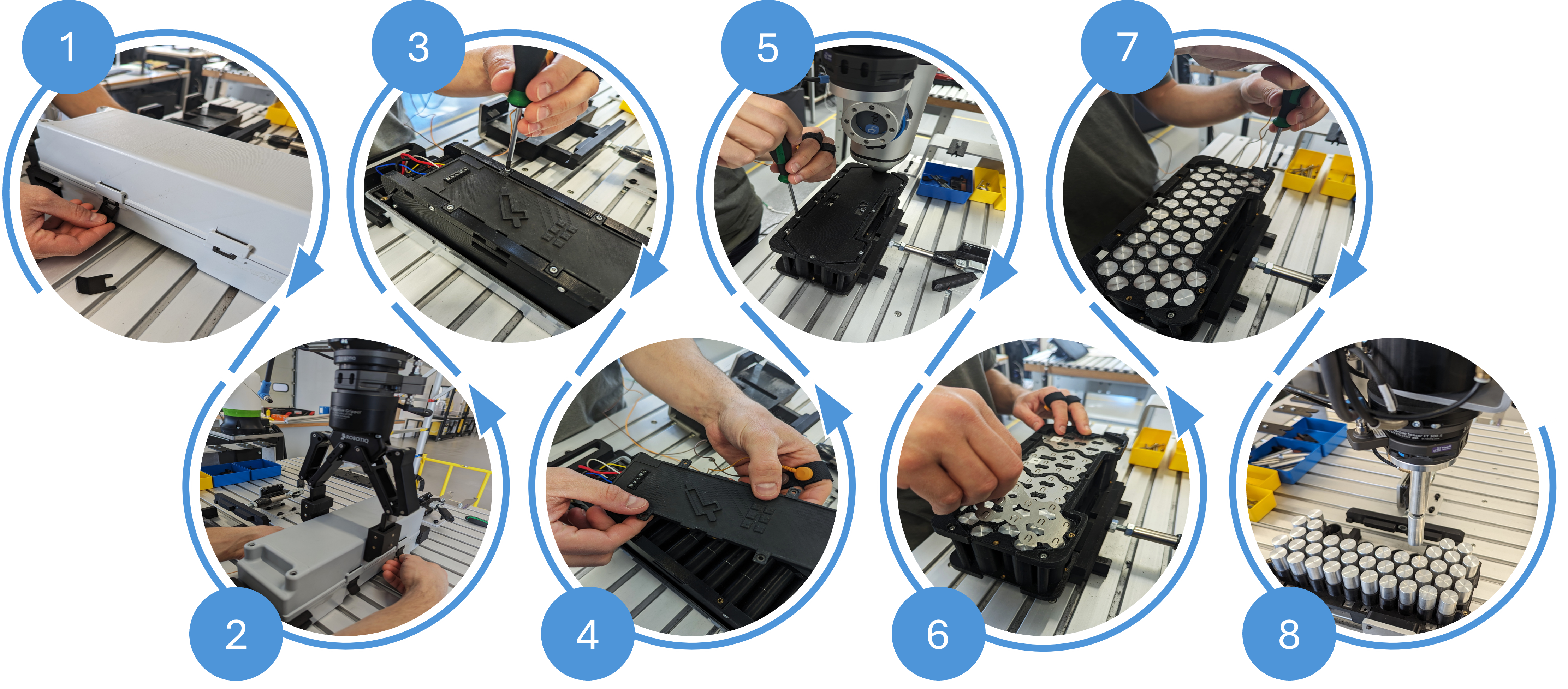}
    \caption{Battery disassembly steps.}
    \label{fig: steps}
\end{figure}

During manual disassembly, the operator opened the battery cover, removed the Battery Management System (BMS), detached the cables, unscrewed the battery components, removed the soldering, and extracted the batteries. In the collaborative disassembly phase, given the difficulty associated with opening the battery casing, this step was conducted collaboratively: the robot pressed against the battery cover to stabilize it, while the human operator loosened the fixturing. Subsequently, while the operator disassembled the BMS, the robot simultaneously unscrewed other battery components. Once the operator finished disassembling the BMS, the human and robot cooperatively unscrewed the remaining components. In Fig. \ref{fig: steps}, the complete set of steps of the collaborative disassembly is represented.

A Fanuc CRX-20\footnote{https://www.fanuc.eu/eu-en/product/robot/crx-20ial} collaborative robot was used for this task. To ensure operator safety, the Fanuc CRX-20 features built-in safety mechanisms, including force and contact sensors, enabling the robot to detect and respond to unexpected physical interactions. The robotic cell used for the data acquisition is shown in Fig. \ref{fig: robotic-cell}. 
The robot was equipped with voice control capabilities, allowing the operator to issue verbal instructions for specific commands. The pipeline consists of an Automatic Speech Recognition (ASR) module and a Natural Language Understanding (NLU) module, which translates the spoken word into robot instructions. This pipeline is presented in \cite{fasana2024}. After receiving the instructions, IPyHOP \cite{Bansod_2022}, a Hierarchical Task Network (HTN) planner, decomposed the high-level command into a sequence of atomic robotic actions. When required, the robot automatically switched tools to execute these actions effectively. The motion trajectories for the robot were computed using the Pilz industrial motion planner from MoveIt2 \cite{coleman2014reducing}, ensuring precise and safe manipulation.

\subsection{Participants}
In total, $55$ subjects participated on the first day of the data collection. The sample mean age is $27.98 \pm 10.22$. $48$ subjects were male and $7$ were female. Out of the $55$, $42$ also participated in the second day. Most subjects were invited from the author's research facility, while the others accepted an external invitation. Participant background varies from undergraduate engineering students to researchers, including professionals in other fields.

\subsection{Ground Truth} \label{sub: ground}
Throughout the experiment, ground truth data were collected by administering multiple self-assessment questionnaires. After each task described in \ref{sec: protocol}, the subjects were asked to answer three questionnaires:
\begin{itemize}
    \item The \textbf{Stress Trait Anxiety Inventory-Y1 (STAI-Y1)} \cite{stai} consists of $20$ questions that measure the subjective feeling of apprehension and worry, and it is often used as a stress measurement.
    \item The \textbf{NASA Task Load Index (NASA-TLX)} \cite{hartDevelopmentNASATLXTask1988} measures self-reported workload and comprises six metrics (mental demand, physical demand, temporal demand, performance, effort, and frustration level).
    \item The \textbf{Self-Assessment Manikin (SAM)} \cite{bradley_1994} assesses participant valence, arousal, and dominance levels. The scale used in this dataset is from one to five.
\end{itemize}
Moreover, at the beginning of the first part of the experiment, participants were asked to complete the Negative Attitude Towards Robots (NARS) \cite{nomuraPsychologyHumanrobotCommunication2004} questionnaire to identify their attitude toward robots.

\subsection{Acquired data}

\begin{figure*}
    \centering
    \includegraphics[width=\linewidth, keepaspectratio]{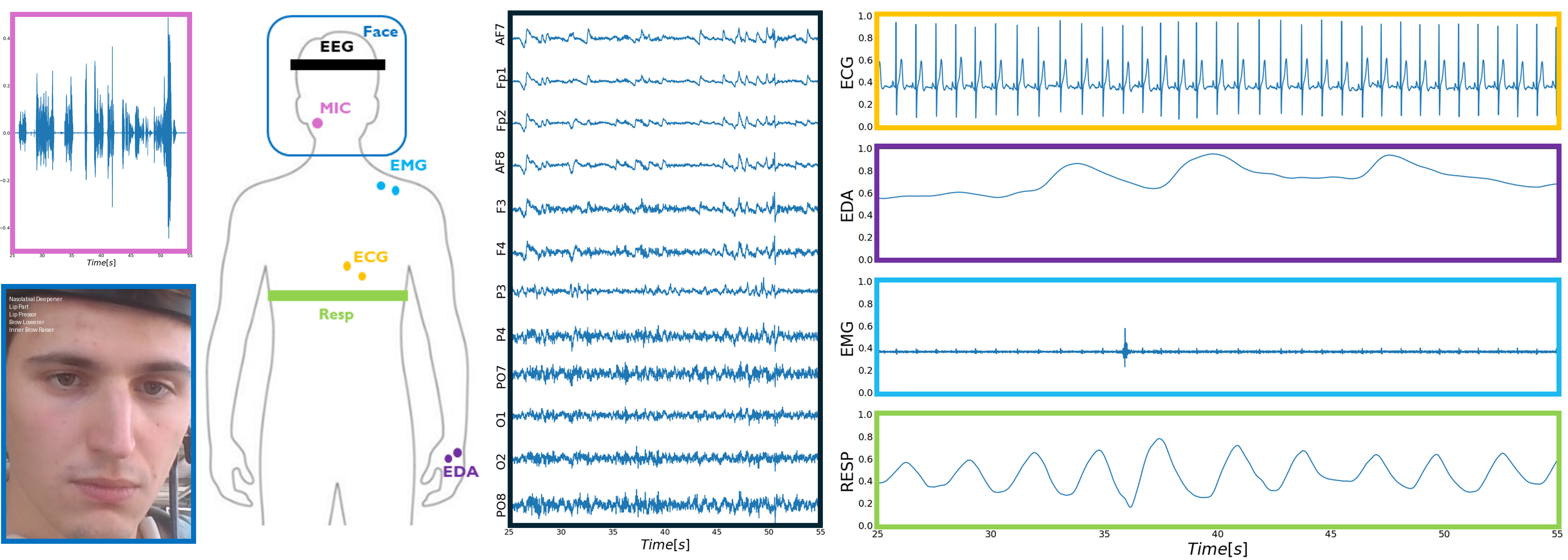}
    \caption{Sample of the acquired physiological data. The participant signals are filtered and normalized (\textit{min-max}).}
    \label{fig: signals}
\end{figure*}

Electroencephalogram signals were acquired using the Bitbrain Diadem\footnote{https://www.bitbrain.com/neurotechnology-products/dry-eeg/diadem}, which is a wearable dry-EEG with 12 sensors over the pre-frontal, frontal, parietal, and occipital brain areas. In particular, the acquired channels are: AF7, Fp1, Fp2, AF8, F3, F4, P3, P4, PO7, O1, O2, PO8, plus ground and reference electrode on the left earlobe.

For the collection of electrocardiogram (ECG), electrodermal activity (EDA), respiration (RESP), and electromyography (EMG), we used the Versatile Bio\footnote{https://www.bitbrain.com/neurotechnology-products/biosignals/versatile-bio} sensor from Bitbrain. The ECG sensor was placed in a V2 configuration to reduce signal noise caused by arm movements. To allow free movement during the experiment, the EDA sensor was placed on the index and middle fingers of the non-dominant hand. The EMG sensor was placed on the right trapezius, while the respiratory band was placed over the subject's chest. In Fig. \ref{fig: signals}, a sample of the collected physiological signals is represented. These devices have been used in other HRC setups such as \cite{loizaga2024}.

Video recordings of the participants were obtained using a standard computer webcam placed in front of the participant during the cognitive tasks and the industrial tasks. Finally, audio recordings were obtained using a commercially available Bluetooth microphone.

All the physiological signals were acquired at $256$ Hz using the software SennsLab\footnote{https://www.bitbrain.com/neurotechnology-products/software/sennslab}. The software manages Bluetooth communication with the devices and synchronizes the physiological signals and audio-video data. The data are displayed in real time, allowing for a visual inspection during the experiment.

\section{Methods} \label{sec: methods}
\subsection{Data processing}
The ECG signals were filtered using a combination of a band-pass filter (with a frequency range from 0.05 to 40 Hz) and a Savitzky–Golay filter. 

Electromyography signals were filtered using a band-pass filter with a frequency range from 10 to 500 Hz coupled with a detrending algorithm, which removes the signal trend by evaluating the linear least-squares fit of the data as specified in the SENIAM recommendations \cite{stegemanStandardsSurfaceElectromyography}. 

The Electrodermal activity signal was filtered using a low-pass filter with a cut-off frequency of 10 Hz, coupled with a convolutional signal smoothing. Then, the signal is down-sampled at 100 Hz and divided into phasic and tonic components using the algorithm presented in \cite{grecoCvxEDAConvexOptimization2016}.

Respiration signals were filtered using a second-order band-pass filter with a frequency range from 0.03 to 5 Hz.

Electroencephalogram signals were processed using two filters: a second-order band-pass filter with a frequency range from 0.5 to 40 Hz and a band-stop filter from 49 to 51 Hz to remove the amplifier noise.

\subsection{Features Extraction} \label{sec: features}
\subsubsection{Physiological data}\label{sub: physio}
Following the processing pipeline, a total of 250 features were extracted from the processed physiological signals, segmented in $60$ seconds windows, using the Neurokit package \cite{Makowski2021neurokit}. These features comprise time-domain, frequency-domain, and complexity measures. For ECG signals, Heart Rate Variability (HRV) features were computed, following the definitions outlined in \cite{phamHeartRateVariability2021}. EMG feature descriptions can be found in \cite{orgucEMGbasedRealTime2018}, while EDA-related features are detailed in \cite{shuklaFeatureExtractionSelection2021}.

Concerning the EEG signals, after processing we segment the signal in $5$ seconds window and compute $7$ for each of the $12$ channels, together with the ratios over the right and left hemispheres ($\theta_{F3}$ / $\alpha_{P3}$, and $\theta_{F4}$ / $\alpha_{P4}$), which where significant to discriminate between levels of mental workload in \cite{raufi2022}. We evaluate the power in the frequency bands ($\gamma$ (30-80 Hz), $\beta$ (13-30 Hz), $\alpha$ (8-13 Hz), $\theta$ (4-8 Hz), and $\delta$ (1-4 Hz)) using Welch's Power Spectral Density (PSD) \cite{welch1967}. Welch’s method estimates the power spectrum of a signal by segmenting it into overlapping windows, computing the Discrete Fourier Transform (DFT) for each window, and then averaging the squared magnitudes. The PSD is computed as follows:

\begin{equation}
    P(\omega) = \frac{1}{K} \sum_{k=1}^{K} \frac{|X_k(\omega)|^2}{M}
\end{equation}

where $X_k(\omega)$ is the DFT of the k-th windowed segment, and M is the number of points in each segment. Moreover, we compute Differential entropy (DiffEn) and Sample Entropy (SampEn) for each channel.

\subsubsection{Face Action Units}\label{sub: AU}
To optimize computational efficiency, facial data were analyzed at a reduced frame rate of 2 fps. Action Unit (AU) detection was performed using the pre-trained XGBoost model from Py-Feat \cite{cheongPyFeatPythonFacial2023}, which identifies the presence of facial muscle activations. The model estimates a probability score for each of the 20 detected action units at every selected frame, forming a multivariate time series per repetition.

\begin{figure}
    \centering
    \includegraphics[width=0.8\linewidth, trim={0.5cm 4cm 2cm 2cm},clip]{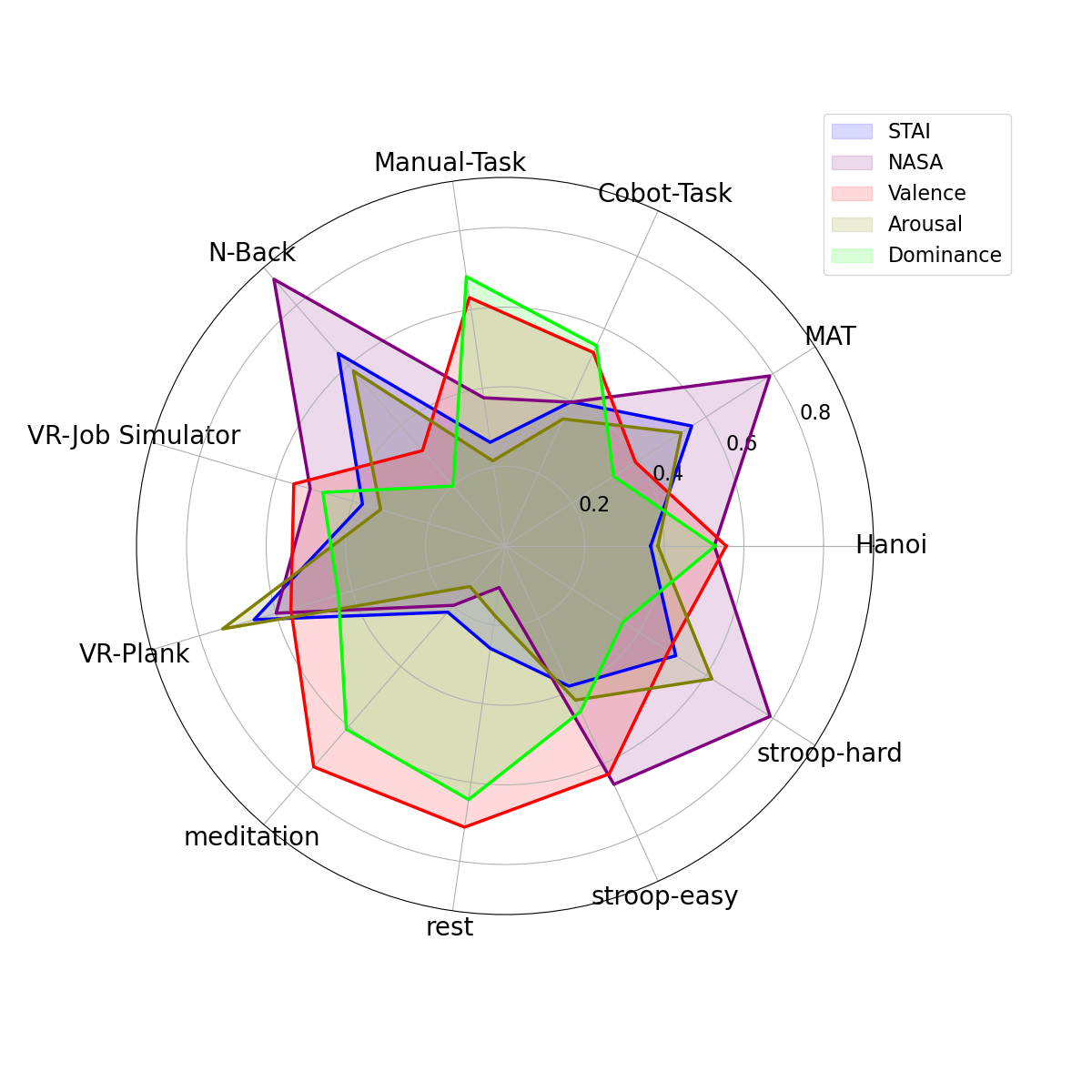}
    \caption{The radar chart displays the mean values of various ground truth metrics (STAI, NASA, Valence, Arousal, and Dominance) across different experimental conditions. The values are normalized (\textit{min-max}) by subject.}
    \label{fig: labels}
\end{figure}

\subsubsection{Voice Features}\label{sub: voice}
The spoken segments were automatically detected using the Silero-VAD model \cite{SileroVAD}. Features consisted of statistical measurements of the fundamental frequency, harmonicity, shimmer, and jitter. Moreover, the features include speech formats and Mel Frequency Cepstrum Coefficients (MFCCs). From the latter, we evaluated statistical measurements such as mean and standard deviation as in \cite{tombaStressDetectionSpeech2018}, but we also included median, kurtosis, and skewness measurements.

\subsubsection{Text embeddings}\label{sub: nlp}
Given the spoken segments, we used the large variant of OpenAI's Whisper model \cite{radford2023robust} to transcribe the voice into text. This transcription is later fed into a Sentence Transformer model \cite{reimers-2019-sentence-bert} to extract the embeddings of the given text. Since participants are all Italian mother tongue, we employed a model fine-tuned for the Italian language \cite{bertITA}.

\begin{table}[t!]
\centering
\renewcommand{\arraystretch}{1.2}
\begin{tabular}{ c | c | c | c }
    Response & Signal & Model & RMSE \\ [1ex]
    \hline \hline
    \multirow{9}{8em}{\centering STAI-Y1 \\ $\mu=31.86$, \\ $max=55.00$} & \multirow{3}{5em}{\centering Physio \\ $n=250$} & RF & $ 0.20 \pm 0.09$ \\
     & & AB & $0.20 \pm 0.09$  \\
     & & XGB & $0.23 \pm 0.09$ \\
     \cline{2-4}
     & \multirow{3}{5em}{\centering EEG \\ $n=88$} & RF & $0.32 \pm 0.08$ \\
     & & AB & $0.30 \pm 0.08$ \\
     & & XGB & $0.32 \pm 0.08$ \\
     \cline{2-4}
     & \multirow{3}{5em}{\centering Voice \\ $n=439$} & RF & $0.32 \pm 0.08$ \\
     & & AB & $0.33 \pm 0.08$ \\
     & & XGB & $0.34 \pm 0.07$ \\
     \hline
     \multirow{9}{8em}{\centering NASA-TLX \\ $\mu=39.56$, \\ $max=91.11$} & \multirow{3}{5em}{\centering Physio \\ $n=250$} & RF & $0.19 \pm 0.08$ \\
     & & AB & $0.19 \pm 0.09$ \\
     & & XGB & $0.20 \pm 0.09$ \\
     \cline{2-4}
     & \multirow{3}{5em}{\centering EEG \\ $n=88$} & RF & $0.31 \pm 0.08$ \\
     & & AB & $0.29 \pm 0.08$ \\
     & & XGB & $0.32 \pm 0.08$ \\
     \cline{2-4}
     & \multirow{3}{5em}{\centering Voice \\ $n=439$} & RF & $0.32 \pm 0.08$ \\
     & & AB & $0.32 \pm 0.08$ \\
     & & XGB & $0.33 \pm 0.08$ \\
\end{tabular}
\caption{Results from the regression of the STAI-Y1 and NASA-TLX scores using baseline models (RF: RandomForest, AB: AdaBoost, XGB: XGBoost)}
\label{tab: regression}
\end{table}

\section{Results} \label{sec: results}
The proposed experimental protocol allows for the identification of a wide range of mental states in the participants. In Fig. \ref{fig: labels}, the average ground truth label for each of the tasks is presented. It can be seen that participants experienced different emotional states and cognitive load during the experiment, allowing the dataset to grasp a more holistic view of the participants' psycho-physical state.

Using the features mentioned in \ref{sec: features}, we assess the performance of out-of-the-box baseline models in a regression and a classification task. As baseline models, we select RandomForest \cite{randomforest2001}, AdaBoost \cite{freund1997}, and XGBoost \cite{chen2016}. To evaluate the baseline models, we performed \textit{Leave-One-Subject-Out} validation and computed the performance as mean and standard deviation across subjects. Both features and labels are normalized (\textit{min-max}) using the maximum and minimum values of each subject. For the sake of simplicity, we evaluated three modalities: the data obtained using the Versatile Bio (ECG, EDA, EMG, RESP), the EEG data, and the voice features.

First, we performed the regression over the normalized scores of NASA and STAI. The results are presented in Tab. \ref{tab: regression}. Here, it can be noticed that physiological data provided the lowest RMSE, suggesting that they carry the most relevant information for estimating stress and cognitive load.

For the classification task, we identified three classes from STAI and NASA-TLX subjects' specific scores collected throughout the entire experience. The \textit{Low} class is identified as the tasks where the subject gave a score lower than $\mu - \delta / 2$, where $\mu$ is the subject's mean score across all the tasks and $\delta$ is the standard deviation. The \textit{Medium} class consists of all tasks where the subject answered with a score between $\mu - \delta / 2$ and $\mu + \delta / 2$. Finally, the tasks with \textit{High} class are the ones where the subject answered with a score higher than $\mu + \delta / 2$. The results for the classification task are presented in Tab. \ref{tab: classification}. In this task, physiological features (ECG, EDA, EMG, RESP) achieved the highest F1 scores, particularly for cognitive load classification.

Overall, physiological signals provide the most informative features for both regression and classification tasks, outperforming EEG and voice-based features. EEG signals contain valuable information but are more susceptible to noise, making their performance slightly lower than physiological data. Voice-based features show the lowest predictive power, suggesting that vocal markers alone may not be sufficient for stress and cognitive load estimation. The results indicate that more advanced machine learning models or multimodal fusion techniques could further enhance predictive performance \cite{buss2024}.

\begin{table}[t!]
\renewcommand{\arraystretch}{1.2}
\begin{tabular}{ c | c | c | c }
    Response & Signal & Model & F1-score \\ [1ex]
    \hline \hline
    \multirow{9}{8em}{\centering Stress Class} & \multirow{3}{5em}{\centering Physio \\ $n=250$} & RF & $0.30 \pm 0.14$ \\
     & & AB & $0.33 \pm  0.14$ \\
     & & XGB & $0.329 \pm 0.14$ \\
     \cline{2-4}
     & \multirow{3}{5em}{\centering EEG \\ $n=88$} & RF & $0.37 \pm 0.12$\\
     & & AB & $0.34 \pm 0.16$ \\
     & & XGB & $0.37 \pm 0.11$ \\
     \cline{2-4}
     & \multirow{3}{5em}{\centering Voice \\ $n=439$} & RF & $0.35 \pm 0.15$ \\
     & & AB & $0.34 \pm 0.12$ \\
     & & XGB & $0.36 \pm 0.12$ \\
     \hline
     \multirow{9}{8em}{\centering Cognitive Load Class} & \multirow{3}{5em}{\centering Physio \\ $n=250$} & RF & $0.39 \pm 0.14$ \\
     & & AB & $0.38 \pm 0.10$ \\
     & & XGB & $0.38 \pm 0.14$  \\
     \cline{2-4}
     & \multirow{3}{5em}{\centering EEG \\ $n=88$} & RF & $0.39 \pm 0.11$ \\
     & & AB & $0.38 \pm 0.18$ \\
     & & XGB & $0.40 \pm 0.13$ \\
     \cline{2-4}
     & \multirow{3}{5em}{\centering Voice \\ $n=439$} & RF & $0.41 \pm 0.15$ \\
     & & AB & $0.37 \pm 0.14$ \\
     & & XGB & $0.38 \pm 0.15$ \\
\end{tabular}
\caption{Results from the classification of the 3 stress classes and of the 3 cognitive load classes using baseline models (RF: RandomForest, AB: AdaBoost, XGB: XGBoost).}
\label{tab: classification}
\end{table}

\section{Conclusion} \label{sec: conclusion}
In this paper, we introduced MultiPhysio-HRC, a multimodal physiological signals dataset for industrial Human-Robot Collaboration (HRC). Our dataset provides a comprehensive collection of physiological signals (EEG, ECG, EDA, RESP, EMG), facial features, and voice data, recorded in multiple scenarios,  including real-world industrial-like settings. Through the diversity of the proposed exercises, we elicited diverse cognitive and emotional states, enabling a rich understanding of human psycho-physical responses.

The baseline models applied to the dataset suggest that physiological signals contain valuable information for estimating cognitive load and stress levels. However, the results indicate that achieving high accuracy remains challenging, underscoring the need for advanced machine learning approaches and multi-modal fusion techniques.

By making MultiPhysio-HRC publicly available, we aim to accelerate research in affective computing and human-aware robotics, fostering safer and more human-centered industrial human-robot collaboration.



\end{document}